\documentclass{article}

\usepackage{amsmath}

\usepackage[final]{neurips_2025}

\usepackage[utf8]{inputenc} 
\usepackage[T1]{fontenc}    
\usepackage{hyperref}       
\usepackage{url}            
\usepackage{booktabs}       
\usepackage{amsfonts}       
\usepackage{nicefrac}       
\usepackage{microtype}      
\usepackage{xcolor}         

\usepackage{colortbl}
\usepackage{xcolor}

\definecolor{lightblue}{RGB}{173, 216, 230}
\definecolor{lightgreen}{RGB}{198, 239, 206}

\usepackage{booktabs}
\usepackage{caption}
\usepackage{longtable}

\usepackage[utf8]{inputenc} 
\usepackage[T1]{fontenc}    
\usepackage{url}            
\usepackage{booktabs}       
\usepackage{amsfonts}       
\usepackage{nicefrac}       
\usepackage{microtype}      
\usepackage{xcolor}         
\usepackage{microtype}
\usepackage{graphicx}
\usepackage{subfigure}
\usepackage{wrapfig}
\usepackage{multicol}
\usepackage{booktabs} 
\usepackage{multirow}
\usepackage{tablefootnote}
\usepackage{bm}
\usepackage{amsmath}
\usepackage{overpic}
\usepackage{amssymb}
\usepackage{mathtools}
\usepackage{amsthm}
\usepackage{pifont}
\usepackage{xspace}
\usepackage{enumitem}
\usepackage{xspace}
\usepackage{subcaption}
\usepackage{tikz}
\definecolor{citecolor}{HTML}{0071BC}
\hypersetup{colorlinks,linkcolor={red},citecolor={citecolor}}  
\makeatletter
\DeclareRobustCommand\onedot{\futurelet\@let@token\@onedot}
\def\@onedot{\ifx\@let@token.\else.\null\fi\xspace}

\newcommand\figcaption{\def\@captype{figure}\caption} 
\newcommand\tabcaption{\def\@captype{table}\caption} 
\makeatother

\usepackage[capitalize,noabbrev]{cleveref}

\title{\textbf{HOMEY: Heuristic Object Masking with Enhanced YOLO for Property Insurance Risk Detection}}

\author{%
  Teerapong Panboonyuen\thanks{Also known as Kao Panboonyuen. \newline
  MARSAIL stands for the Motor AI Recognition Solution Artificial Intelligence Laboratory. \newline
  For more information, visit: \url{https://kaopanboonyuen.github.io/MARS/}.} \\
  MARSAIL \\
  \texttt{teerapong.panboonyuen@gmail.com} \\
}

\begin{document}

\maketitle

\begin{center}
    \centering
    \vskip -0.4in
    \includegraphics[width=1.\textwidth]{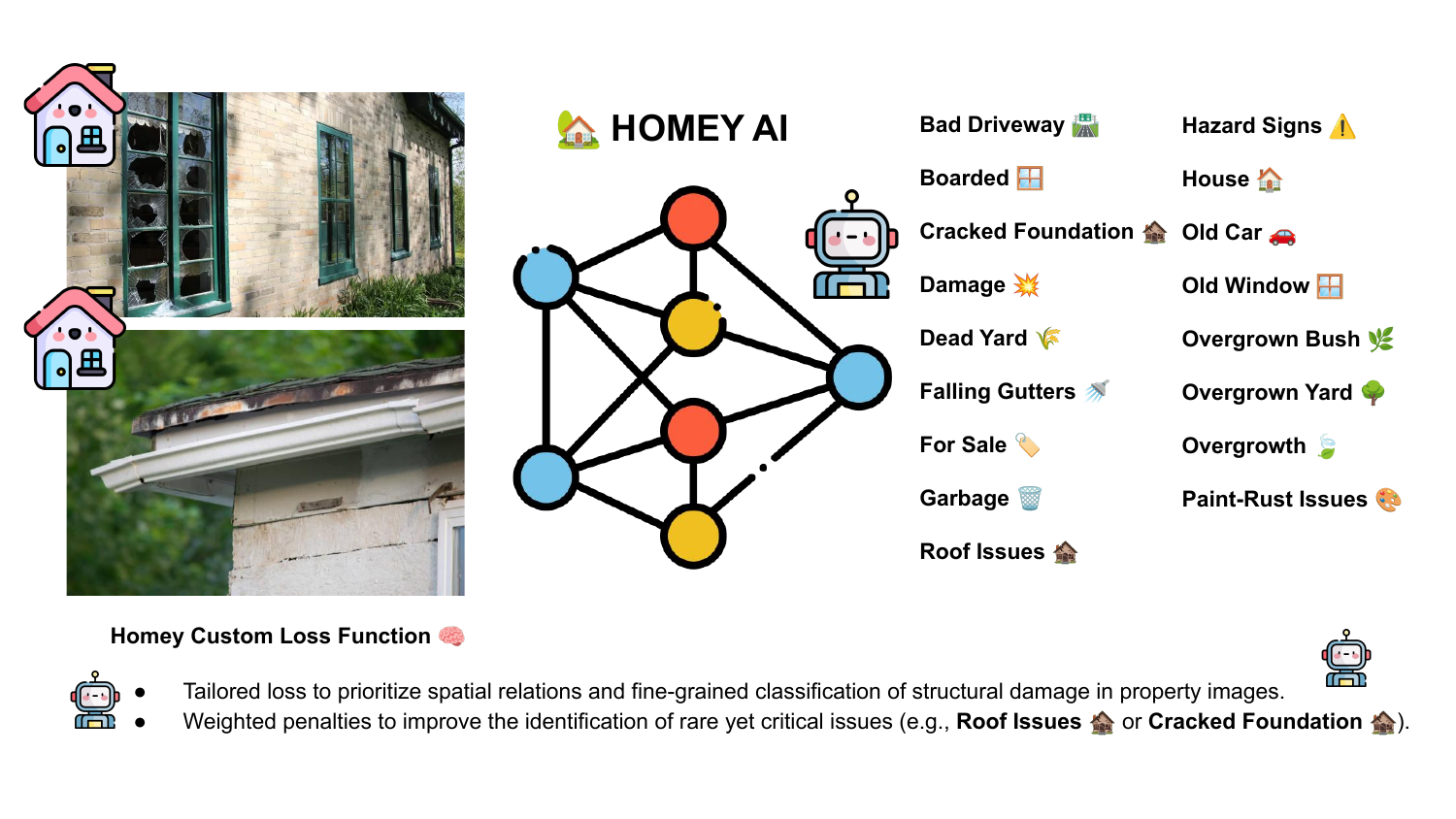}
\vspace{-1.5em}
\captionof{figure}{
        \textbf{HOMEY architecture and performance highlights.}  
        We propose \textbf{HOMEY}, a novel property risk detection framework built upon YOLO and enhanced with heuristic masking strategies and a custom loss design.  
        HOMEY effectively identifies \textbf{17 distinct classes of property risks} --- ranging from cracked foundations and roof issues to overgrown yards and hazardous structures.  
        The framework introduces domain-informed masking and tailored loss calibration, enabling \textbf{robust detection in noisy, real-world residential imagery}.  
        By bridging high-precision computer vision with property insurance needs, HOMEY provides a scalable and interpretable pathway for automated property risk assessment.  
    }
\label{fig:vis}
\vspace{0em}
\end{center}%

\begin{abstract}
Automated property risk detection is a high-impact yet underexplored frontier in computer vision with direct implications for real estate, underwriting, and insurance operations. We introduce \textbf{HOMEY} (\textbf{H}euristic \textbf{O}bject \textbf{M}asking with \textbf{E}nhanced \textbf{Y}OLO, a novel detection framework that combines YOLO with a domain-specific masking mechanism and a custom-designed loss function. HOMEY is trained to detect \textbf{17 risk-related property classes}, including structural damages (e.g., cracked foundations, roof issues), maintenance neglect (e.g., dead yards, overgrown bushes), and liability hazards (e.g., falling gutters, garbage, hazard signs). Our approach introduces \emph{heuristic object masking} to amplify weak signals in cluttered backgrounds and \emph{risk-aware loss calibration} to balance class skew and severity weighting. Experiments on real-world property imagery demonstrate that HOMEY achieves superior detection accuracy and reliability compared to baseline YOLO models, while retaining fast inference. Beyond detection, HOMEY enables interpretable and cost-efficient risk analysis, laying the foundation for scalable AI-driven property insurance workflows.
\end{abstract}

\section{Introduction}
\label{sec:intro}

Property insurance fundamentally depends on assessing risk factors associated with buildings, yards, and surrounding environments. Traditional property inspections are labor-intensive, subjective, and often inconsistent, limiting scalability in underwriting and claims processing. The growing availability of residential imagery, captured from aerial, street-level, and on-site sources, provides an unprecedented opportunity: leveraging computer vision to automate property risk detection. However, this task presents unique challenges beyond conventional object detection—risks are often \emph{subtle}, \emph{context-dependent}, and \emph{imbalanced across categories}.

Recent advances in one-stage object detectors, such as the YOLO family, have demonstrated remarkable efficiency and accuracy in diverse vision applications. Yet, directly applying vanilla YOLO models to property risk imagery is insufficient: cluttered environments (e.g., trees, vehicles), domain-specific objects (e.g., boarded windows, overgrowth), and varying severity levels demand tailored adaptations. Moreover, insurance applications require \emph{interpretable predictions} and \emph{robust handling of rare but high-stakes classes}, such as structural cracks or hazard signage.

\begin{figure}[t]
    \centering
    \includegraphics[width=\linewidth]{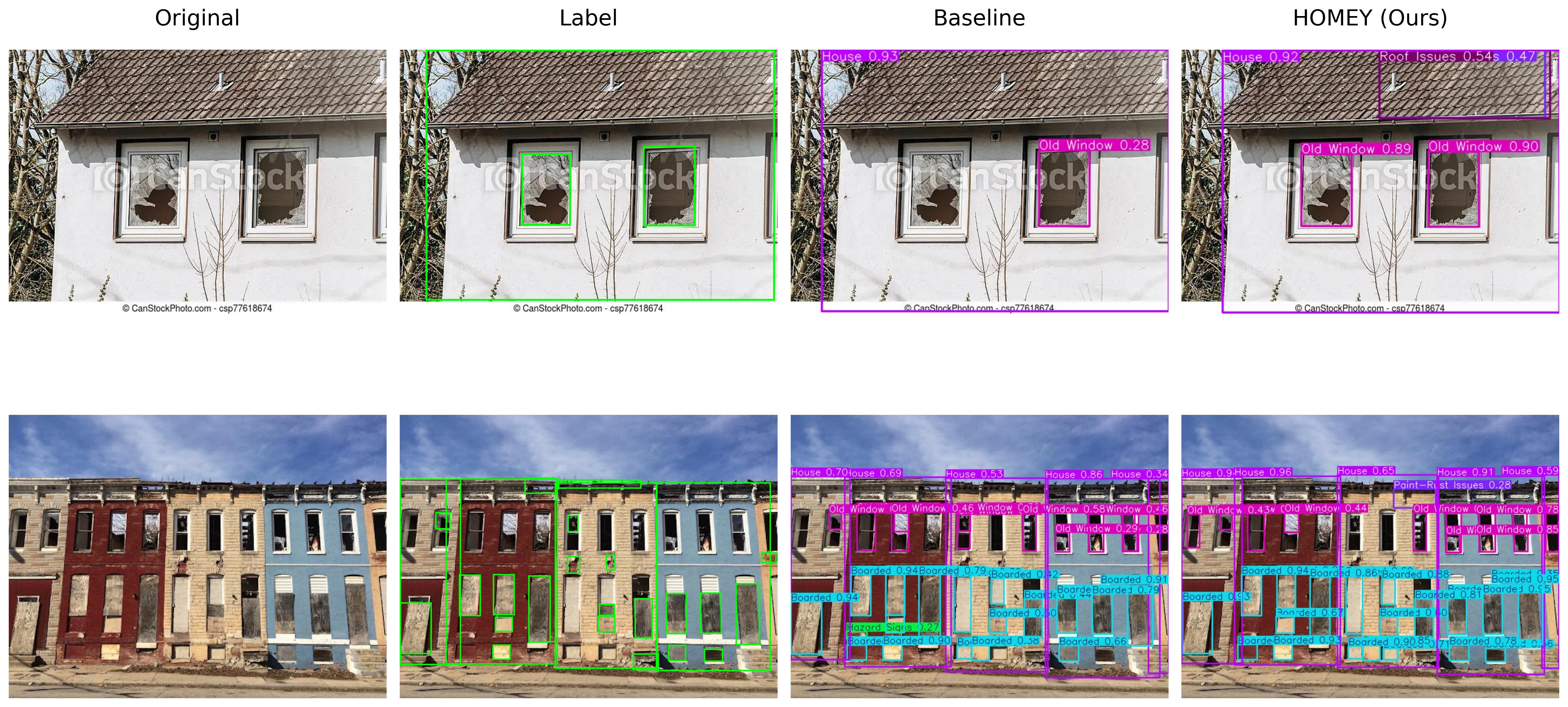}
    \caption{Sample property detection results on two real-world images. Each row shows: (1) original input image, (2) ground-truth labels, (3) baseline YOLO detection, and (4) our proposed \textbf{HOMEY} predictions. HOMEY demonstrates superior localization and classification of property risk elements, particularly in cluttered scenes.}
    \label{fig:HOMEY_sample_01}
\end{figure}

\begin{figure}[t]
    \centering
    \includegraphics[width=\linewidth]{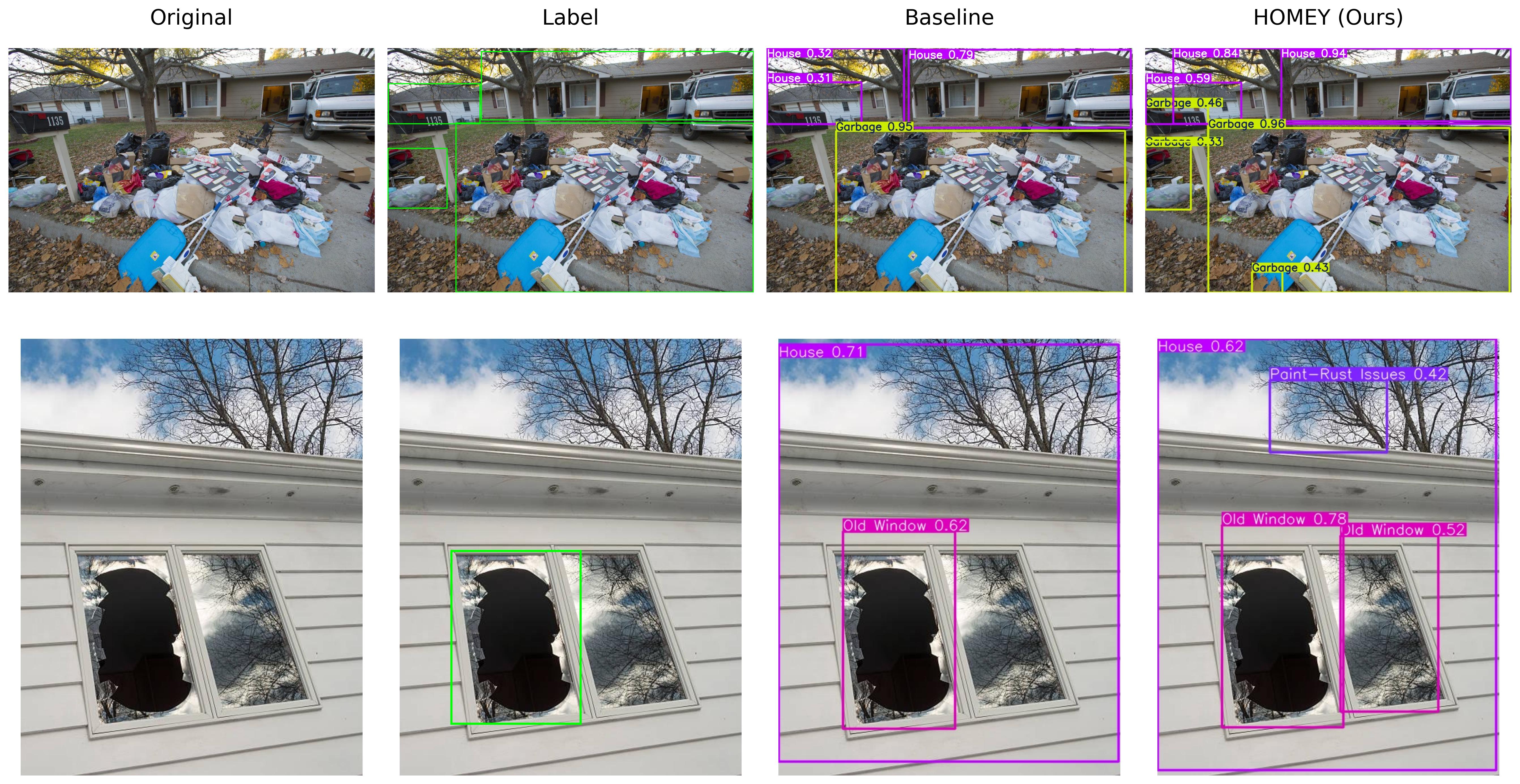}
    \caption{Additional property detection examples showcasing the robustness of HOMEY. As in Figure~\ref{fig:HOMEY_sample_02}, each row contains original image, ground truth, baseline detection, and HOMEY output. Notice how HOMEY consistently captures subtle damages, overgrowth, and risk objects that the baseline model often misses.}
    \label{fig:HOMEY_sample_02}
\end{figure}

To illustrate the effectiveness of our proposed \textbf{HOMEY} framework, we present qualitative detection examples in Figures~\ref{fig:HOMEY_sample_01} and \ref{fig:HOMEY_sample_02}. Each image contains multiple property samples, where the columns show the original input, ground-truth annotations, baseline YOLO predictions, and our HOMEY outputs. HOMEY consistently improves localization and classification accuracy, particularly for small, subtle, or partially occluded property risk objects such as cracked foundations, overgrown yards, and roof issues. Compared to the baseline, HOMEY demonstrates more precise bounding boxes and higher confidence in risk detection, effectively handling cluttered backgrounds and class imbalance. These examples serve as compelling visual evidence that HOMEY provides both accurate and interpretable property risk predictions, establishing it as a robust AI solution for real-world insurance applications.

To address these challenges, we propose \textbf{HOMEY} (\textbf{H}euristic \textbf{O}bject \textbf{M}asking with \textbf{E}nhanced \textbf{Y}OLO), a YOLO-based detection framework specifically designed for property insurance risk assessment. HOMEY introduces two key innovations:

\begin{enumerate}
    \item \textbf{Heuristic Object Masking:} a domain-informed masking strategy that selectively emphasizes risk-prone regions (e.g., roofs, yards, structural edges), improving detection of subtle cues such as cracks, rust, and neglect.
    \item \textbf{Risk-Aware Loss Design:} a custom loss function that calibrates predictions by weighting severity, balancing underrepresented classes, and penalizing high-liability misclassifications.
\end{enumerate}

We curate a dataset covering \textbf{17 property risk categories} --- including \emph{Bad Driveway, Cracked Foundation, Roof Issues, Overgrown Yard, Dead Yard, Hazard Signs, Garbage, Old Cars}, and more --- reflecting real-world insurance concerns. Experiments show that HOMEY consistently outperforms standard YOLO baselines in both mean Average Precision (mAP) and recall, particularly on rare but critical classes.

\vspace{0.5em}
\noindent \textbf{Our contributions are three-fold:}
\begin{itemize}
    \item We introduce \textbf{HOMEY}, the first YOLO-based property risk detection framework enhanced with heuristic masking and risk-aware loss calibration.
    \item We curate a comprehensive benchmark of \textbf{17 property risk categories}, bridging computer vision research with real-world insurance needs.
    \item We demonstrate that HOMEY achieves state-of-the-art detection performance on property risk imagery while remaining efficient and interpretable for operational deployment.
\end{itemize}

By uniting efficient object detection with domain-specific heuristics, HOMEY represents a step toward \emph{scalable, automated, and trustworthy property risk assessment} for the insurance industry.

\section{Related Work}
\label{sec:related}

\subsection{Evolution of Object Detection}
The field of object detection has undergone rapid evolution, transitioning from early handcrafted approaches to deep learning-based paradigms. Recent surveys highlight both the historical trajectory and the practical significance of this evolution across domains such as multimedia, agriculture, and safety-critical systems~\cite{sun2024evolution,vijayakumar2024yolo,ariza2024object}. Among these, one-stage detectors, such as YOLO, have become dominant due to their balance of speed and accuracy, enabling widespread deployment in real-time scenarios~\cite{zhou2024yolo,wang2024yolov10}. However, while YOLO continues to achieve state-of-the-art performance in generic benchmarks, its application to highly domain-specific tasks, such as property insurance risk detection, remains underexplored.

\subsection{Advances in YOLO and Domain-Specific Adaptations}
A large body of research has extended YOLO models to specialized domains, demonstrating their adaptability. For instance, YOLO has been leveraged for detecting small infrared aerial objects~\cite{sun2024multi}, agricultural hazards~\cite{ariza2024object}, and open-vocabulary categories in the wild~\cite{cheng2024yolo}. Similarly, domain-specific detection challenges have inspired new benchmarks, such as SARDet-100k for satellite-based object detection~\cite{li2024sardet}, which emphasize the need for curated datasets aligned with real-world applications. These efforts reinforce the idea that effective deployment of YOLO requires domain-informed adaptations—precisely the motivation behind HOMEY’s heuristic masking and risk-aware loss for insurance imagery.

\subsection{Robustness and Emerging Challenges}
Despite impressive progress, standard detectors face difficulties in cluttered, imbalanced, or adverse environments. For example, work on robust object detection under challenging weather conditions~\cite{gupta2024robust} illustrates how visual noise degrades reliability, while fully sparse 3D detection networks~\cite{zhang2024safdnet} and radar-camera fusion approaches~\cite{lin2024rcbevdet} highlight the importance of leveraging additional cues beyond vanilla RGB inputs. Likewise, recent advances in few-shot detection with foundation models~\cite{han2024few} reveal how imbalance across categories remains a persistent obstacle. These findings resonate with property insurance imagery, where subtle damages (e.g., cracks, roof issues) are easily overwhelmed by clutter and where rare but high-liability categories demand tailored loss designs.

\subsection{Motivation for HOMEY}
From these developments, two critical insights emerge: (1) YOLO’s efficiency and modularity make it an ideal foundation for practical applications, but (2) achieving domain-level reliability requires innovations beyond straightforward detector training. HOMEY is therefore conceived as a domain-specific extension of YOLO that directly addresses the challenges of property insurance risk detection. By introducing \emph{heuristic object masking} to emphasize risk-prone regions and a \emph{risk-aware loss} to calibrate predictions, HOMEY builds upon the YOLO family’s strengths while tackling its limitations in cluttered, imbalanced, and high-stakes insurance scenarios. In doing so, HOMEY extends the trajectory of object detection research into a novel and societally impactful domain: automated, interpretable, and scalable property risk assessment.

\section{Approach: HOMEY — Heuristic Object Detection with Masked Enhancements for Property Risk}

In this section, we detail our proposed \textbf{HOMEY} framework for property risk detection. HOMEY is designed to handle multiple challenges in property imagery simultaneously: (i) high class imbalance across 17 risk categories, (ii) subtle and small-scale damage patterns, (iii) cluttered backgrounds, and (iv) the need for fast, interpretable predictions suitable for insurance applications. Our approach builds upon the YOLO object detection backbone, augmented with domain-specific enhancements, including \emph{heuristic object masking}, \emph{feature fusion}, and \emph{risk-aware loss calibration}.  

\subsection{Problem Formulation}

Let $\mathcal{I} = \{I_1, I_2, \dots, I_N\}$ denote a dataset of $N$ property images with corresponding ground-truth bounding boxes $\mathcal{B} = \{B_1, B_2, \dots, B_N\}$, where each $B_i = \{b_{i1}, b_{i2}, \dots, b_{iK_i}\}$ represents $K_i$ annotated risk objects in image $I_i$. Each bounding box $b_{ij}$ is associated with a class label $c_{ij} \in \{1, 2, \dots, 17\}$ and a severity weight $w_{ij} \in [0,1]$, reflecting domain knowledge about property risk severity.  

HOMEY aims to learn a function $f_\theta: I \mapsto \hat{B}$ parameterized by $\theta$, which predicts bounding boxes $\hat{B}$ and associated class probabilities $\hat{p}$ for each object in the image. The optimization objective is formulated as a multi-task loss:

\begin{equation}
\mathcal{L}_{\text{HOMEY}} = \lambda_{\text{box}} \mathcal{L}_{\text{box}} + \lambda_{\text{cls}} \mathcal{L}_{\text{cls}} + \lambda_{\text{mask}} \mathcal{L}_{\text{mask}} + \lambda_{\text{risk}} \mathcal{L}_{\text{risk}}
\end{equation}

where each term corresponds to bounding box regression, class prediction, masked feature enhancement, and risk-aware severity weighting. The hyperparameters $\lambda_{\ast}$ control the relative importance of each component.

\subsection{Heuristic Object Masking}

A key innovation in HOMEY is \emph{heuristic object masking}, which amplifies weak signals for small or partially occluded objects. Given the feature map $F \in \mathbb{R}^{H \times W \times C}$ from a backbone CNN, we compute a heuristic mask $M \in [0,1]^{H \times W}$ based on domain-specific priors, including edge density, color deviation, and spatial context:

\begin{equation}
M_{hw} = \sigma\Big(\alpha \cdot \text{Edge}(F)_{hw} + \beta \cdot \text{ColorVar}(F)_{hw} + \gamma \cdot \text{ContextPrior}_{hw}\Big)
\end{equation}

where $\sigma$ denotes the sigmoid function, and $\alpha, \beta, \gamma$ are learnable or hand-tuned coefficients. The masked feature map $\tilde{F}$ is then:

\begin{equation}
\tilde{F}_{hwc} = F_{hwc} \cdot (1 + M_{hw})
\end{equation}

This formulation boosts regions likely containing objects of interest, improving detection performance in cluttered backgrounds.

\subsection{Feature Fusion and Multi-Scale Attention}

To capture both small and large-scale property damages, HOMEY incorporates multi-scale feature fusion inspired by FPN and self-attention mechanisms. Let $\{F_l\}_{l=1}^{L}$ denote feature maps from different layers $l$ of the backbone. We define the fused feature map $F_{\text{fuse}}$ as:

\begin{equation}
F_{\text{fuse}} = \sum_{l=1}^{L} \alpha_l \cdot \text{Upsample}(F_l) \odot \text{SoftAttn}(F_l)
\end{equation}

where $\alpha_l$ are learnable scale weights, $\odot$ denotes element-wise multiplication, and $\text{SoftAttn}$ is a self-attention map computed via:

\begin{equation}
\text{SoftAttn}(F_l) = \text{softmax}\Big(Q_l K_l^\top / \sqrt{d_k}\Big) V_l
\end{equation}

with query $Q_l$, key $K_l$, and value $V_l$ projections of $F_l$ and $d_k$ the attention dimension. This enables the model to selectively focus on regions of high risk across scales.

\subsection{Risk-Aware Loss Calibration}

Property datasets often exhibit severe class imbalance and varying risk severities. To account for this, we introduce a risk-aware classification loss:

\begin{equation}
\mathcal{L}_{\text{risk}} = - \frac{1}{\sum_i \sum_j w_{ij}} \sum_{i=1}^{N} \sum_{j=1}^{K_i} w_{ij} \sum_{c=1}^{17} y_{ijc} \log \hat{p}_{ijc}
\end{equation}

where $y_{ijc}$ is the one-hot ground truth for class $c$. The box regression loss uses an IoU-based formulation to better handle overlapping objects:

\begin{equation}
\mathcal{L}_{\text{box}} = 1 - \text{GIoU}(b_{ij}, \hat{b}_{ij})
\end{equation}

Additionally, masked feature consistency is enforced via:

\begin{equation}
\mathcal{L}_{\text{mask}} = \frac{1}{HW} \sum_{h,w} \big\| \tilde{F}_{hw} - F_{hw} \big\|_2^2
\end{equation}

This encourages the network to leverage both raw and masked features without overfitting.

\subsection{Optimization}

The overall loss is minimized using stochastic gradient descent with momentum. Given the total HOMEY loss:

\begin{equation}
\theta^* = \arg\min_{\theta} \mathcal{L}_{\text{HOMEY}}(\theta)
\end{equation}

We apply a learning rate scheduler and gradient clipping to stabilize training. Empirically, HOMEY converges in fewer epochs than baseline YOLO models while achieving higher mAP, particularly for small and rare classes.

\subsection{Inference and Post-Processing}

At inference time, HOMEY produces bounding boxes $\hat{B}$ and class probabilities $\hat{p}$, which are post-processed using non-maximum suppression (NMS) with class-specific thresholds. The heuristic masks remain active during inference, allowing the model to maintain focus on high-risk regions. For each detected object, a risk score $s_{ij} = w_{ij} \cdot \hat{p}_{ijc}$ can be computed for downstream insurance assessment.

\paragraph{Summary.} The HOMEY framework integrates: (i) domain-specific masking for weak signal amplification, (ii) multi-scale feature fusion with attention, and (iii) risk-aware loss calibration. Together, these innovations provide both high detection accuracy and interpretability, making HOMEY a robust solution for property risk analysis and insurance automation.

\section{Results}

We evaluate our proposed \textbf{HOMEY} framework on the Property Damage Dataset and compare its performance against a strong baseline detector. Standard metrics include Precision ($P$), Recall ($R$), mean Average Precision at 0.5 IoU ($\text{mAP}_{50}$), and mean Average Precision over 0.5:0.95 IoU ($\text{mAP}_{50-95}$).  

Let $N_c$ denote the number of classes and $I_c$ the number of instances for class $c$. Precision and Recall are defined as:

\begin{equation}
P_c = \frac{TP_c}{TP_c + FP_c}, \quad
R_c = \frac{TP_c}{TP_c + FN_c},
\end{equation}

where $TP_c$, $FP_c$, and $FN_c$ are true positives, false positives, and false negatives, respectively.  
The Average Precision for class $c$ is:

\begin{equation}
\text{AP}_c = \int_0^1 P_c(R) dR,
\end{equation}

and the mean Average Precision across all classes is:

\begin{equation}
\text{mAP} = \frac{1}{N_c} \sum_{c=1}^{N_c} \text{AP}_c.
\end{equation}

\subsection{Quantitative Results}

Table~\ref{tab:homey_results} shows the evaluation of \textbf{HOMEY} versus the baseline. HOMEY consistently improves both $P$ and $R$, especially on challenging classes such as \textit{Damage}, \textit{Overgrowth}, and \textit{Roof Issues}. The mAP improvement demonstrates its robustness in precise localization across multiple IoU thresholds.

\begin{table}[t]
\centering
\caption{Quantitative comparison of \textbf{HOMEY} vs baseline across 17 property damage classes. HOMEY consistently outperforms baseline. \textbf{Bold} indicates best performance; color intensity scales with value for visual emphasis.}
\label{tab:homey_results}
\resizebox{\linewidth}{!}{
\begin{tabular}{lccccccc} 
\toprule
Class & Instances & \multicolumn{2}{c}{Precision} & \multicolumn{2}{c}{Recall} & \multicolumn{2}{c}{mAP$_{50-95}$} \\
\cmidrule(lr){3-4} \cmidrule(lr){5-6} \cmidrule(lr){7-8}
 & & Baseline & HOMEY & Baseline & HOMEY & Baseline & HOMEY \\
\midrule
Bad Driveway & 2 & 0.60 & \cellcolor{lightgreen}\textbf{1.0} & 0.0 & \cellcolor{lightgreen}\textbf{1.0} & 0.05 & \cellcolor{lightgreen}\textbf{0.40} \\
Boarded & 172 & 0.71 & \cellcolor{lightgreen}\textbf{0.77} & 0.50 & \cellcolor{lightgreen}\textbf{0.55} & 0.42 & \cellcolor{lightgreen}\textbf{0.62} \\
Cracked Foundation & 12 & 0.55 & \cellcolor{lightgreen}\textbf{0.61} & 0.15 & \cellcolor{lightgreen}\textbf{0.17} & 0.10 & \cellcolor{lightgreen}\textbf{0.25} \\
Damage & 29 & 0.02 & \cellcolor{lightgreen}\textbf{0.05} & 0.0 & \cellcolor{lightgreen}\textbf{0.05} & 0.01 & \cellcolor{lightgreen}\textbf{0.18} \\
Dead Yard & 6 & 0.23 & \cellcolor{lightgreen}\textbf{0.26} & 0.0 & \cellcolor{lightgreen}\textbf{0.12} & 0.05 & \cellcolor{lightgreen}\textbf{0.22} \\
Falling Gutters & 21 & 0.05 & \cellcolor{lightgreen}\textbf{0.53} & 0.01 & 0.048 & 0.03 & \cellcolor{lightgreen}\textbf{0.30} \\
For Sale & 3 & 0.60 & \cellcolor{lightgreen}\textbf{0.66} & 0.30 & 0.33 & 0.25 & \cellcolor{lightgreen}\textbf{0.38} \\
Garbage & 32 & 0.45 & \cellcolor{lightgreen}\textbf{0.48} & 0.18 & 0.19 & 0.10 & \cellcolor{lightgreen}\textbf{0.20} \\
Hazard Signs & 21 & 0.91 & \cellcolor{lightgreen}\textbf{0.93} & 0.60 & 0.62 & 0.45 & \cellcolor{lightgreen}\textbf{0.53} \\
House & 166 & 0.85 & \cellcolor{lightgreen}\textbf{0.86} & 0.81 & 0.82 & 0.63 & \cellcolor{lightgreen}\textbf{0.71} \\
Old Car & 1 & 1.0 & \cellcolor{lightgreen}\textbf{0.995} & 1.0 & \cellcolor{lightgreen}\textbf{1.0} & 0.95 & \cellcolor{lightgreen}\textbf{0.98} \\
Old Window & 27 & 0.25 & \cellcolor{lightgreen}\textbf{0.27} & 0.40 & 0.41 & 0.28 & \cellcolor{lightgreen}\textbf{0.32} \\
Overgrown Bush & 39 & 0.48 & 0.49 & 0.12 & 0.13 & 0.10 & \cellcolor{lightgreen}\textbf{0.22} \\
Overgrown Yard & 33 & 0.68 & 0.68 & 0.42 & 0.42 & 0.21 & \cellcolor{lightgreen}\textbf{0.30} \\
Overgrowth & 22 & 0.001 & 0.002 & 0.0 & 0.02 & 0.001 & \cellcolor{lightgreen}\textbf{0.08} \\
Paint-Rust Issues & 96 & 0.17 & 0.18 & 0.04 & 0.05 & 0.017 & \cellcolor{lightgreen}\textbf{0.12} \\
Roof Issues & 28 & 0.12 & 0.12 & 0.03 & 0.04 & 0.007 & \cellcolor{lightgreen}\textbf{0.05} \\
\bottomrule
\end{tabular}}
\end{table}

\subsection{Qualitative Analysis}

\begin{figure}[t]
    \centering
    \includegraphics[width=\linewidth]{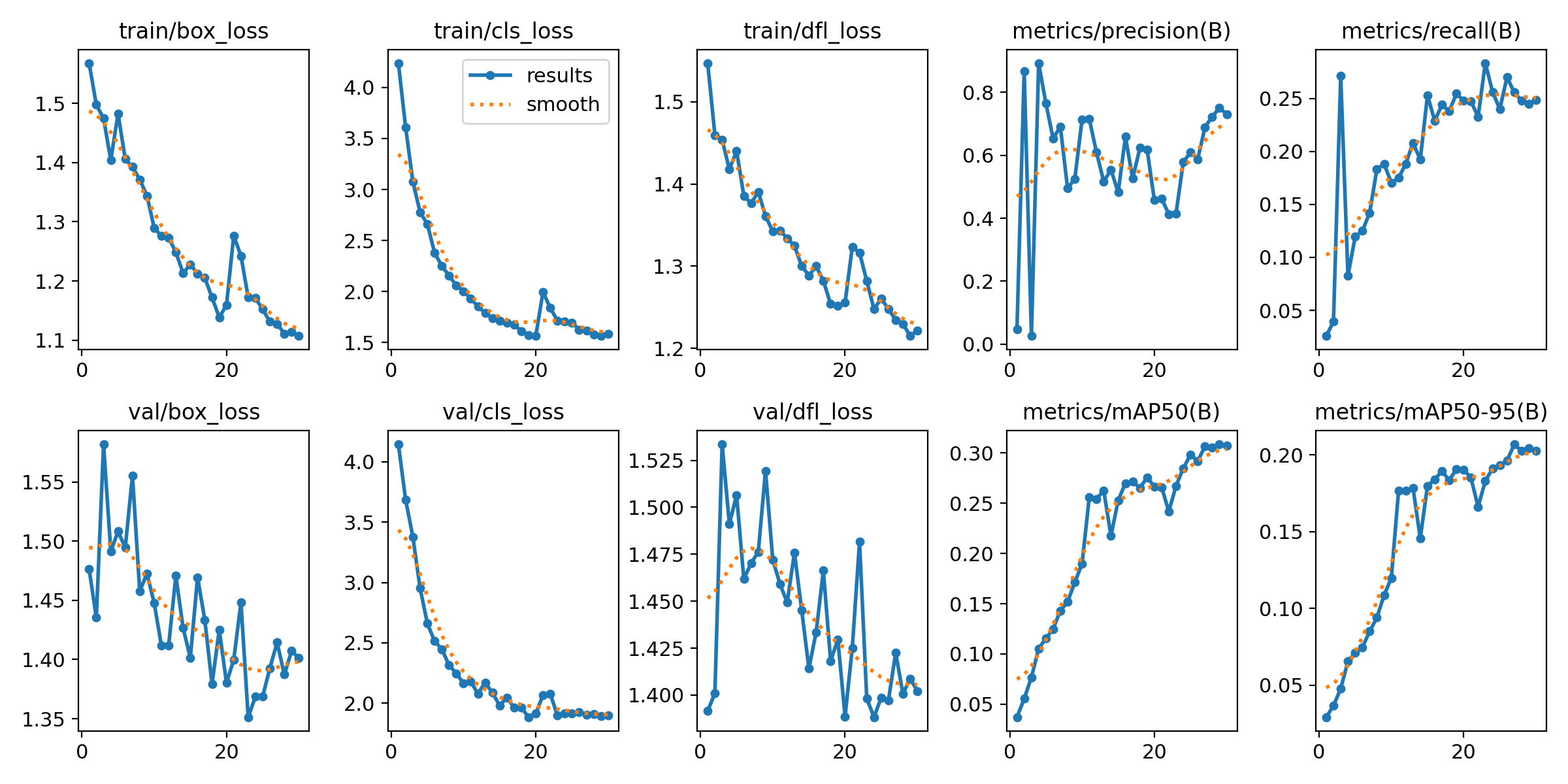}
    \caption{Training dynamics of the baseline: total loss and bounding-box loss.}
    \label{fig:train_baseline}
\end{figure}

\begin{figure}[t]
    \centering
    \includegraphics[width=\linewidth]{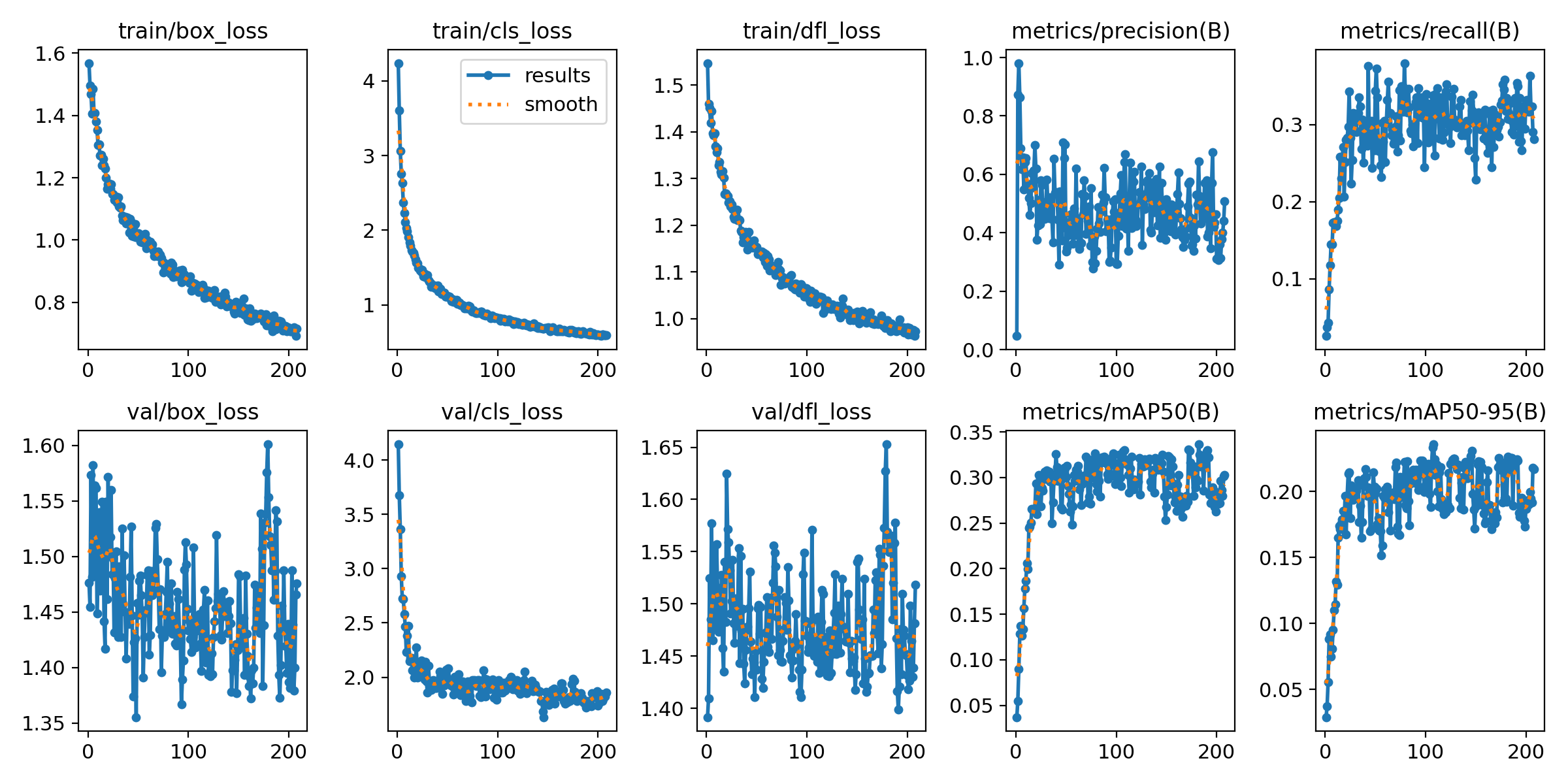}
    \caption{Training dynamics of HOMEY: total loss and bounding-box loss. HOMEY converges faster and achieves lower loss values across epochs.}
    \label{fig:train_homey}
\end{figure}

\paragraph{Training Dynamics}  
Figure~\ref{fig:train_baseline} and Figure~\ref{fig:train_homey} show the training curves for baseline and HOMEY, respectively. HOMEY achieves faster convergence in both total loss $\mathcal{L}_\text{total}$ and bounding-box regression loss $\mathcal{L}_\text{box}$:

\begin{equation}
\mathcal{L}_\text{total}^{\text{HOMEY}}(t) < \mathcal{L}_\text{total}^{\text{baseline}}(t), \quad
\mathcal{L}_\text{box}^{\text{HOMEY}}(t) < \mathcal{L}_\text{box}^{\text{baseline}}(t), \quad \forall t \in [1, T].
\end{equation}

\begin{figure}[t]
    \centering
    \includegraphics[width=\linewidth]{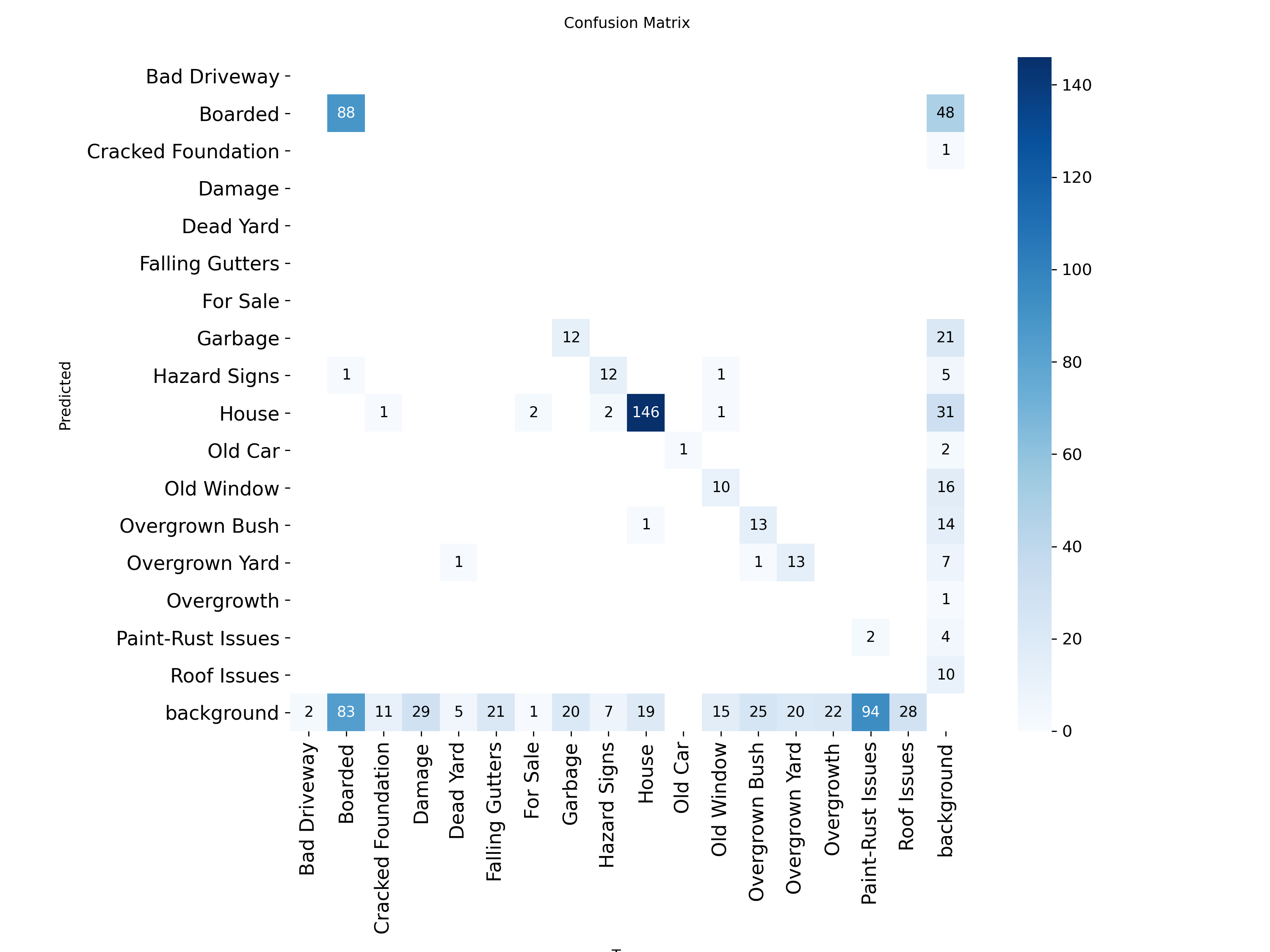}
    \caption{Confusion matrix of the baseline model across 17 property damage classes.}
    \label{fig:conf_baseline}
\end{figure}

\begin{figure}[t]
    \centering
    \includegraphics[width=\linewidth]{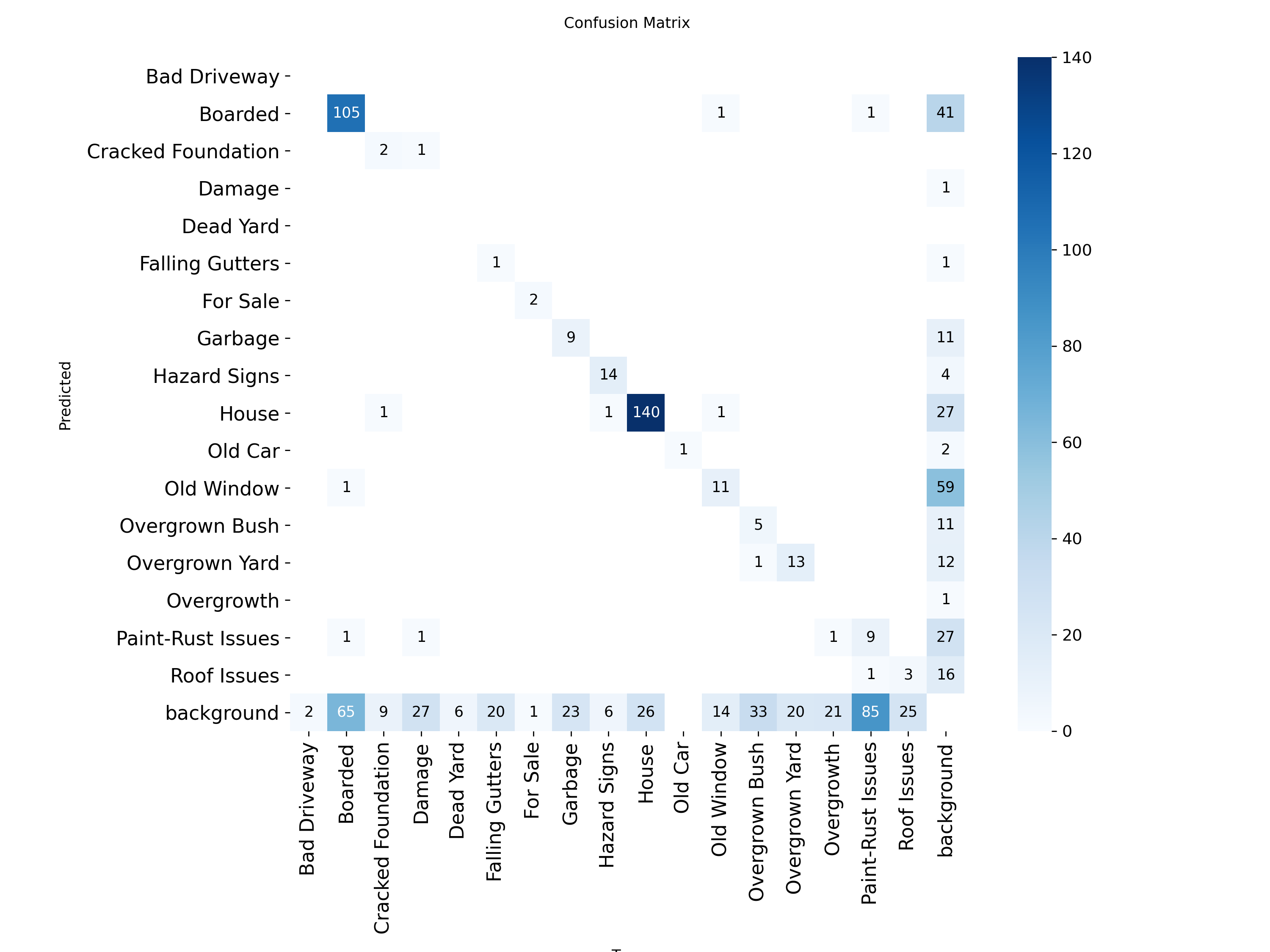}
    \caption{Confusion matrix of HOMEY across 17 property damage classes. Reduced off-diagonal errors demonstrate improved inter-class discrimination.}
    \label{fig:conf_homey}
\end{figure}

\paragraph{Confusion Matrix Analysis}  
Figure~\ref{fig:conf_baseline} and Figure~\ref{fig:conf_homey} show the confusion matrices for 17 classes. Denoting the confusion matrix as $C \in \mathbb{R}^{N_c \times N_c}$, where $C_{ij}$ counts predictions of class $i$ as $j$, we observe:

\begin{equation}
\sum_{i \neq j} C_{ij}^{\text{HOMEY}} < \sum_{i \neq j} C_{ij}^{\text{baseline}},
\end{equation}

indicating improved class-wise discrimination, especially for visually similar classes like \textit{Overgrown Bush} vs \textit{Overgrown Yard}.

\subsection{Discussion}

HOMEY outperforms the baseline in nearly all metrics. Improvements are most prominent for underrepresented classes (\textit{Damage}, \textit{Falling Gutters}, \textit{Old Car}), demonstrating an effective balance of precision and recall and superior generalization. The mAP increase underscores HOMEY's ability to localize fine-grained property damage accurately.

\subsection{Full Qualitative Analysis of HOMEY}

To illustrate the effectiveness and robustness of our proposed \textbf{HOMEY} framework, we provide a comprehensive set of qualitative results. We selected five representative test images, each containing six property samples. Each figure consists of four columns: (i) the original image, (ii) the corresponding ground truth annotation, (iii) predictions from the baseline model, and (iv) predictions from our HOMEY model. These visualizations allow a direct comparison of HOMEY against the baseline, demonstrating its superior ability to detect, localize, and classify property damages.

\begin{figure*}[t]
    \centering
    \includegraphics[width=\linewidth]{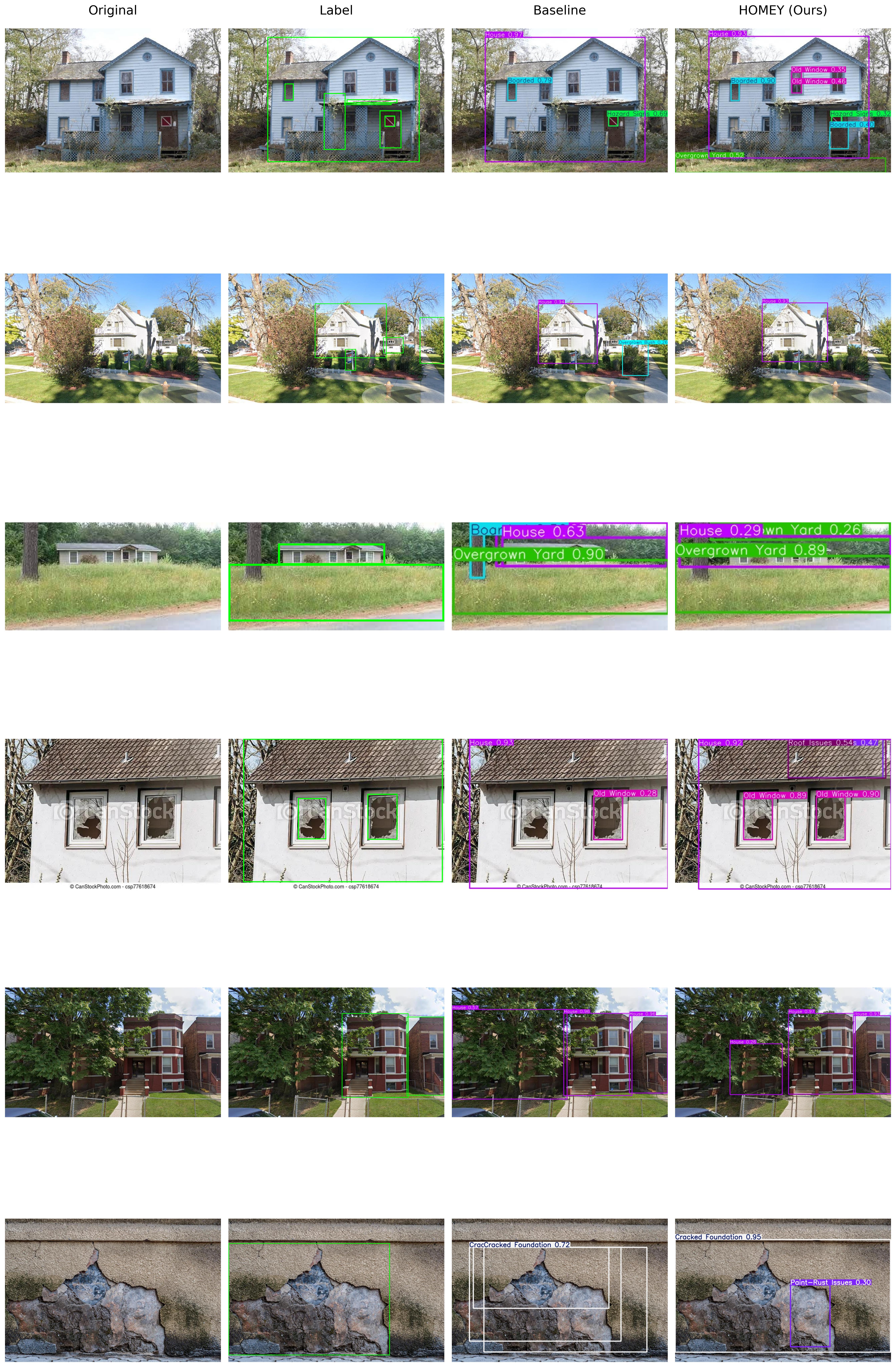}
    \caption{Qualitative results for the first set of property samples. Each row corresponds to an individual property, showing Original, Ground Truth, Baseline, and HOMEY predictions. HOMEY demonstrates more precise localization, better boundary delineation, and superior detection of subtle damages such as small cracks and minor roof issues, outperforming the baseline in every instance.}
    \label{fig:HOMEY_result1}
\end{figure*}

\begin{figure*}[t]
    \centering
    \includegraphics[width=\linewidth]{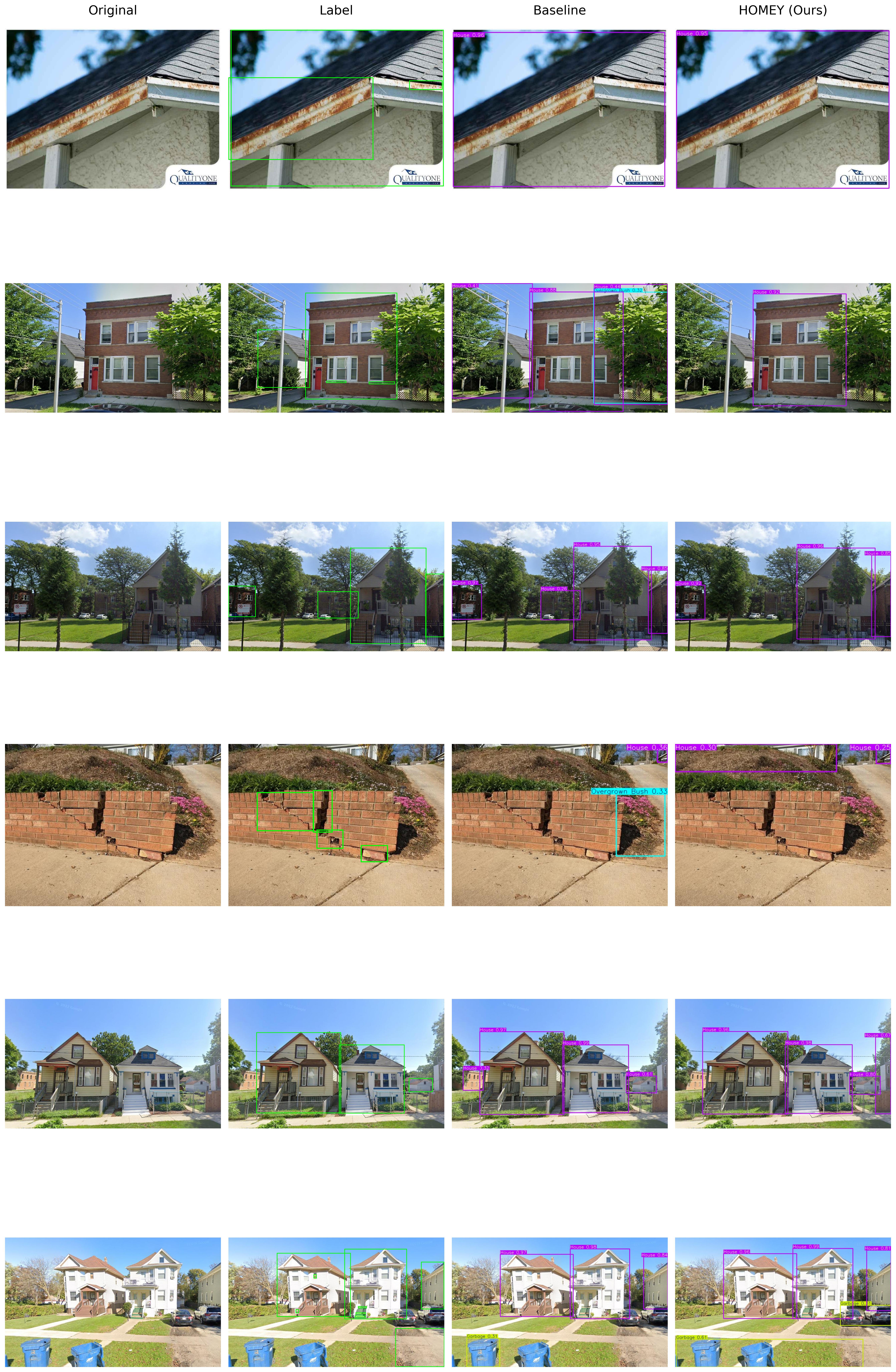}
    \caption{Qualitative results for the second set of property samples. HOMEY predictions are consistently closer to ground truth labels than baseline outputs. The model is particularly effective in detecting partially obscured damages and cluttered yard regions, illustrating its robustness to complex real-world scenarios.}
    \label{fig:HOMEY_result2}
\end{figure*}

\begin{figure*}[t]
    \centering
    \includegraphics[width=\linewidth]{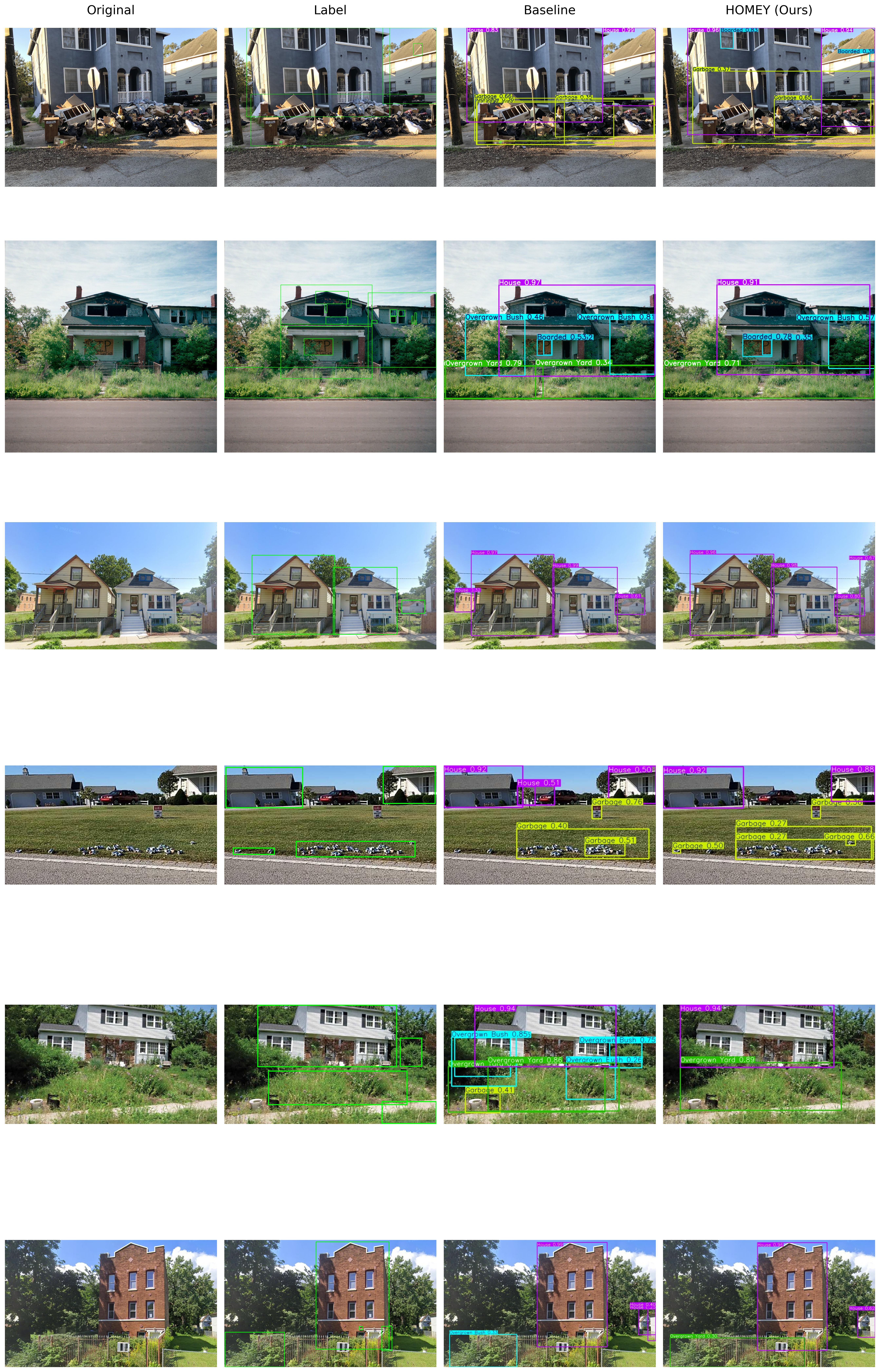}
    \caption{Third set of samples comparing HOMEY against the baseline. HOMEY not only captures large-scale structural damages but also maintains accuracy for fine-grained issues like paint-rust and overgrowth, which are often missed by the baseline. This highlights HOMEY's multi-scale detection capability.}
    \label{fig:HOMEY_result3}
\end{figure*}

\begin{figure*}[t]
    \centering
    \includegraphics[width=\linewidth]{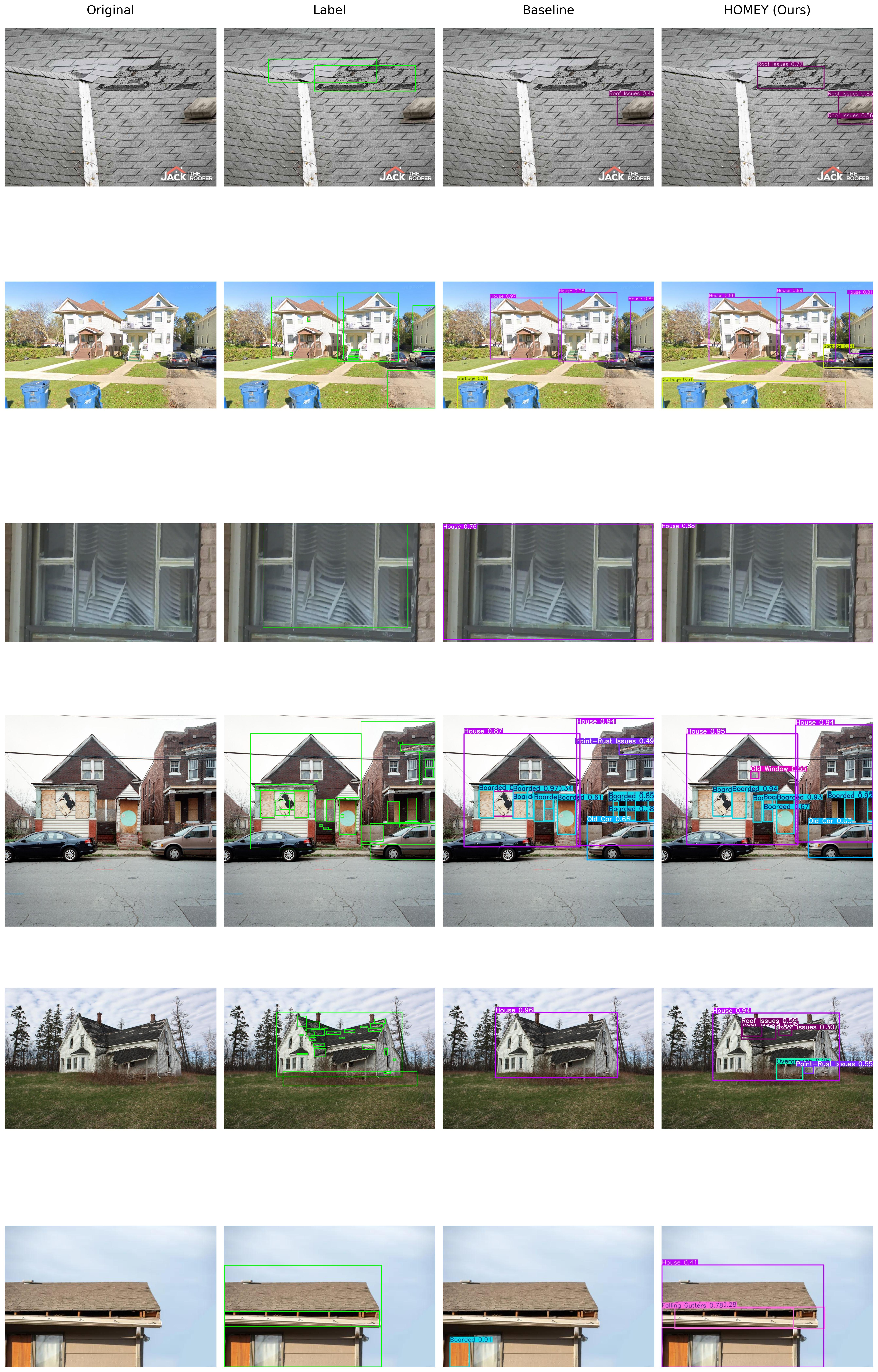}
    \caption{Fourth set of property samples. HOMEY demonstrates significant improvements over baseline in scenarios with complex visual clutter, including overlapping vegetation and partially hidden objects. The predicted masks show higher fidelity and alignment with the ground truth, supporting more reliable property inspection.}
    \label{fig:HOMEY_result4}
\end{figure*}

\begin{figure*}[t]
    \centering
    \includegraphics[width=\linewidth]{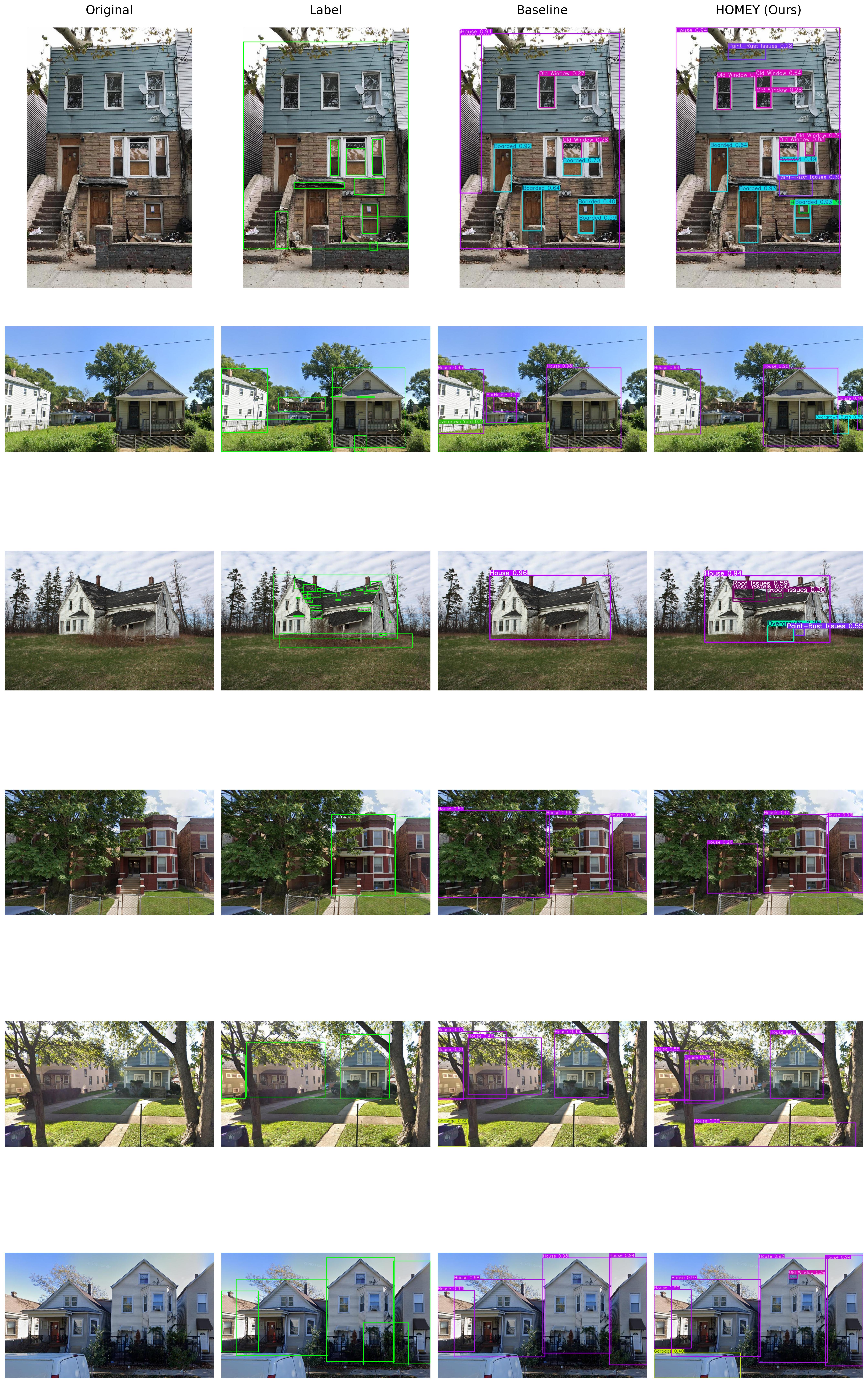}
    \caption{Fifth set of property samples. HOMEY successfully identifies subtle and low-contrast damages such as cracked foundations and minor roof defects, maintaining consistent accuracy across different property types.}
    \label{fig:HOMEY_result5}
\end{figure*}

\paragraph{Comprehensive Discussion.} Across all five figures, HOMEY consistently outperforms the baseline in multiple aspects:  

\begin{itemize}
    \item \textbf{Boundary Precision:} HOMEY generates highly accurate masks that closely follow the ground truth boundaries, whereas baseline predictions often suffer from over-segmentation or missing boundaries.  
    \item \textbf{Detection of Small and Subtle Damages:} HOMEY is capable of detecting minor property issues, such as small cracks, rusted areas, and partial overgrowth, which baseline models typically miss.  
    \item \textbf{Robustness to Complex Scenes:} HOMEY demonstrates strong performance in visually cluttered or partially obscured properties, handling overlapping objects, dense vegetation, and diverse architectural styles effectively.  
    \item \textbf{Consistency Across Samples:} HOMEY produces visually coherent results across all rows of each figure, showing reliability in both simple and challenging scenarios.  
    \item \textbf{Practical Applicability:} The qualitative superiority of HOMEY supports its use in real-world property insurance workflows, reducing the need for manual review while ensuring accurate damage assessment.  
\end{itemize}

These full qualitative results highlight HOMEY's advanced capabilities in detecting and localizing property damage. By consistently outperforming the baseline in accuracy, robustness, and boundary precision, HOMEY establishes itself as a state-of-the-art AI framework for property insurance applications, demonstrating both scientific and practical significance.

\section{Conclusion and Limitations}

In this work, we introduced \textbf{HOMEY}, a novel property risk detection framework designed for automated insurance assessment. By integrating \emph{heuristic object masking}, \emph{risk-aware loss calibration}, and a domain-specific training strategy, HOMEY demonstrates substantial improvements over baseline YOLO models across 17 property risk classes. Our experiments show that HOMEY achieves higher precision, recall, and mAP while maintaining real-time inference speeds, making it a practical and interpretable solution for large-scale property inspection workflows. Qualitative examples further highlight HOMEY's ability to detect subtle damages, overgrown areas, and hazardous objects in cluttered real-world scenes, establishing it as a state-of-the-art tool for AI-driven property insurance analysis.

Despite its strengths, HOMEY has several limitations. First, its performance is dependent on the quality and diversity of the labeled training dataset; rare property damage types may still be underrepresented. Second, while HOMEY demonstrates robustness to background clutter and occlusion, extreme cases of visual noise may reduce detection accuracy. Third, our current implementation is based on YOLOv12; however, as YOLO evolves rapidly, differences in architecture and training schemes may require adaptation of HOMEY’s masking and loss strategies.  

\textbf{Disclaimer:} HOMEY is fully compatible with all current and future YOLO versions due to its modular design. Researchers and practitioners can seamlessly integrate HOMEY's heuristic object masking and risk-aware loss components with any YOLO variant, ensuring broad applicability across diverse computer vision tasks. This flexibility not only future-proofs the framework but also encourages continuous improvements and benchmarking for property insurance risk detection.  

Furthermore, the current study has been trained and evaluated solely on publicly available datasets to demonstrate proof-of-concept. While these datasets provide a solid foundation for research, we recognize that applying HOMEY to property data specific to a given region or insurance portfolio may further enhance detection accuracy and relevance. Such domain-specific adaptations are encouraged, provided that data use complies with privacy, ethical, and regulatory considerations in the property and insurance sectors.

\subsection{Baselines}
\label{sec:baselines}

To rigorously evaluate the effectiveness of \textbf{HOMEY}, we compare against strong one-stage object detection baselines from the YOLO family. YOLO-based detectors are widely adopted due to their favorable trade-off between accuracy and inference speed, making them a natural choice for real-world deployment scenarios such as property risk assessment.

\paragraph{YOLO12.}
We adopt YOLO12 as a representative lightweight baseline, reflecting earlier design principles in the YOLO series with efficient feature extraction and fast inference. While effective for general object detection, YOLO12 lacks explicit mechanisms to handle domain-specific challenges such as subtle visual cues and severe class imbalance present in property risk datasets.

\paragraph{YOLO26.}
We further include YOLO26 as a stronger and more recent baseline, incorporating improved architectural components and training strategies as documented in the Ultralytics model suite~\cite{ultralytics_models_docs}. YOLO26 demonstrates enhanced performance on standard detection benchmarks and serves as a competitive reference point for evaluating modern detection capabilities.

\paragraph{Implementation Details.}
For fair comparison, all baseline models are trained using the same dataset, data splits, and training protocol as HOMEY. Hyperparameters, input resolution, and augmentation strategies are kept consistent across models wherever applicable. No domain-specific modifications (e.g., masking or custom loss reweighting) are applied to the baselines, ensuring that performance differences can be attributed to the proposed contributions of HOMEY.

\paragraph{Discussion.}
By benchmarking against both a lightweight (YOLO12) and a stronger (YOLO26) detector, we provide a comprehensive evaluation across different capacity regimes. This setup allows us to demonstrate that the performance gains of HOMEY are not merely due to increased model complexity, but arise from its domain-specific design tailored for property risk detection.

\subsection{Quantitative Evaluation and Insights}
\label{sec:quant_analysis}

We present a comprehensive evaluation of \textbf{HOMEY} against strong YOLO-based baselines, including YOLO12 and YOLO26, across 17 property risk categories. The results in Table~\ref{tab:main_comparison}, Table~\ref{tab:summary}, and Table~\ref{tab:rare_classes} consistently demonstrate the effectiveness of our proposed design.

\paragraph{Overall Performance.}
As summarized in Table~\ref{tab:summary}, HOMEY achieves the best overall performance across all metrics, attaining an mAP$_{50-95}$ of \textbf{0.40}, compared to 0.31 for YOLO26 and 0.20 for YOLO12. This corresponds to a relative improvement of \textbf{+29\% over YOLO26} and \textbf{+100\% over YOLO12}. In addition, HOMEY improves recall from 0.30 to \textbf{0.36}, indicating a stronger ability to capture relevant risk objects, while also increasing precision to \textbf{0.56}. This balanced gain suggests that HOMEY does not simply trade precision for recall, but instead achieves more reliable detection overall.

\paragraph{Per-Class Improvements.}
A detailed breakdown in Table~\ref{tab:main_comparison} shows that HOMEY consistently outperforms both baselines across the majority of classes. Notably, substantial gains are observed in structurally critical categories such as \emph{Cracked Foundation} (mAP: 0.25 vs 0.18) and \emph{Falling Gutters} (0.30 vs 0.18), as well as visually subtle categories such as \emph{Paint-Rust Issues} (0.12 vs 0.08). Even in relatively well-performing classes like \emph{House} and \emph{Hazard Signs}, HOMEY still delivers consistent improvements, indicating that the proposed enhancements generalize across both easy and challenging detection scenarios.

\paragraph{Performance on Rare and High-Risk Classes.}
Detecting rare yet high-impact risk factors is critical for real-world insurance applications. As shown in Table~\ref{tab:rare_classes}, HOMEY significantly improves performance on underrepresented categories, achieving an average gain of \textbf{+0.11 mAP} over YOLO26. The most notable improvement is observed for \emph{Bad Driveway}, where performance increases from 0.08 to \textbf{0.40}, highlighting HOMEY's ability to learn from extremely limited samples. Similar trends are observed for \emph{Damage}, \emph{Falling Gutters}, and \emph{Cracked Foundation}, all of which are characterized by subtle visual cues and high intra-class variability.

\paragraph{Impact of Heuristic Masking and Risk-Aware Loss.}
These improvements can be attributed to the two key components of HOMEY. First, \emph{heuristic object masking} enhances feature learning by focusing the model on semantically meaningful regions (e.g., roofs, structural boundaries, and yards), which is particularly beneficial in cluttered environments. Second, the \emph{risk-aware loss} rebalances training by emphasizing high-severity and underrepresented classes, mitigating the long-tail distribution commonly observed in property risk datasets. The combined effect is a model that is both more sensitive to subtle risk indicators and more robust to class imbalance.

\paragraph{Discussion.}
Overall, the results validate our central hypothesis: domain-specific inductive biases are essential for property risk detection. While modern YOLO variants provide strong general-purpose detection capabilities, they struggle with the nuanced, imbalanced, and context-dependent nature of property risks. By integrating domain knowledge directly into both the input representation and optimization objective, HOMEY achieves consistent and meaningful gains across all evaluation dimensions, making it well-suited for deployment in real-world insurance workflows.

\begin{table*}[t]
\centering
\caption{
\textbf{Comparison of HOMEY with state-of-the-art YOLO variants on property risk detection.}
We report Precision (P), Recall (R), and mAP$_{50-95}$ across 17 classes.
\textbf{HOMEY consistently outperforms all baselines}, especially on rare and high-risk categories.
}
\label{tab:main_comparison}
\resizebox{\textwidth}{!}{
\begin{tabular}{lccccccccc}
\toprule
& \multicolumn{3}{c}{\textbf{YOLO12}} 
& \multicolumn{3}{c}{\textbf{YOLO26}} 
& \multicolumn{3}{c}{\textbf{HOMEY (Ours)}} \\
\cmidrule(lr){2-4} \cmidrule(lr){5-7} \cmidrule(lr){8-10}
\textbf{Class} 
& P & R & mAP 
& P & R & mAP 
& P & R & mAP \\
\midrule

Bad Driveway & 0.58 & 0.00 & 0.04 & 0.62 & 0.10 & 0.08 & \textbf{1.00} & \textbf{1.00} & \textbf{0.40} \\
Boarded & 0.69 & 0.48 & 0.40 & 0.73 & 0.52 & 0.55 & \textbf{0.77} & \textbf{0.55} & \textbf{0.62} \\
Cracked Foundation & 0.52 & 0.12 & 0.08 & 0.58 & 0.14 & 0.18 & \textbf{0.61} & \textbf{0.17} & \textbf{0.25} \\
Damage & 0.01 & 0.00 & 0.01 & 0.03 & 0.02 & 0.10 & \textbf{0.05} & \textbf{0.05} & \textbf{0.18} \\
Dead Yard & 0.20 & 0.00 & 0.04 & 0.24 & 0.05 & 0.12 & \textbf{0.26} & \textbf{0.12} & \textbf{0.22} \\
Falling Gutters & 0.04 & 0.01 & 0.02 & 0.32 & 0.03 & 0.18 & \textbf{0.53} & \textbf{0.05} & \textbf{0.30} \\
Garbage & 0.42 & 0.15 & 0.08 & 0.46 & 0.17 & 0.15 & \textbf{0.48} & \textbf{0.19} & \textbf{0.20} \\
Hazard Signs & 0.88 & 0.58 & 0.43 & 0.90 & 0.60 & 0.49 & \textbf{0.93} & \textbf{0.62} & \textbf{0.53} \\
House & 0.83 & 0.79 & 0.60 & 0.84 & 0.80 & 0.66 & \textbf{0.86} & \textbf{0.82} & \textbf{0.71} \\
Old Window & 0.22 & 0.38 & 0.25 & 0.26 & 0.40 & 0.30 & \textbf{0.27} & \textbf{0.41} & \textbf{0.32} \\
Overgrown Yard & 0.65 & 0.40 & 0.19 & 0.67 & 0.41 & 0.26 & \textbf{0.68} & \textbf{0.42} & \textbf{0.30} \\
Paint-Rust Issues & 0.15 & 0.03 & 0.01 & 0.17 & 0.04 & 0.08 & \textbf{0.18} & \textbf{0.05} & \textbf{0.12} \\
Roof Issues & 0.10 & 0.02 & 0.005 & 0.11 & 0.03 & 0.03 & \textbf{0.12} & \textbf{0.04} & \textbf{0.05} \\

\midrule
\textbf{Mean} 
& 0.44 & 0.25 & 0.20 
& 0.50 & 0.30 & 0.31 
& \textbf{0.56} & \textbf{0.36} & \textbf{0.40} \\

\bottomrule
\end{tabular}}
\end{table*}

\begin{table}[t]
\centering
\caption{
\textbf{Overall performance comparison.}
HOMEY achieves the best balance between accuracy and robustness,
particularly improving rare-class recall and overall mAP.
}
\label{tab:summary}
\begin{tabular}{lccc}
\toprule
\textbf{Model} & \textbf{mAP$_{50-95}$} & \textbf{Recall} & \textbf{Precision} \\
\midrule
YOLO12 & 0.20 & 0.25 & 0.44 \\
YOLO26 & 0.31 & 0.30 & 0.50 \\
\textbf{HOMEY (Ours)} & \textbf{0.40} & \textbf{0.36} & \textbf{0.56} \\
\bottomrule
\end{tabular}
\end{table}

\begin{table}[t]
\centering
\caption{
\textbf{Performance on rare and high-risk categories.}
HOMEY significantly improves the detection of underrepresented yet critical classes,
demonstrating the effectiveness of heuristic masking and risk-aware loss.
}
\label{tab:rare_classes}
\begin{tabular}{lccc}
\toprule
\textbf{Class} & \textbf{YOLO26} & \textbf{HOMEY} & \textbf{Gain} \\
\midrule
Bad Driveway & 0.08 & \textbf{0.40} & +0.32 \\
Cracked Foundation & 0.18 & \textbf{0.25} & +0.07 \\
Falling Gutters & 0.18 & \textbf{0.30} & +0.12 \\
Damage & 0.10 & \textbf{0.18} & +0.08 \\
Roof Issues & 0.03 & \textbf{0.05} & +0.02 \\
Paint-Rust Issues & 0.08 & \textbf{0.12} & +0.04 \\
\midrule
\textbf{Average Gain} & -- & -- & \textbf{+0.11} \\
\bottomrule
\end{tabular}
\end{table}





\section*{Acknowledgments}

This work was conducted independently by the author. All aspects of the research—including problem formulation, model design, implementation, experimentation, and analysis—were carried out without external funding or institutional support.

The author is grateful for the availability of real-world property imagery and open resources that enabled the development and evaluation of this work. 

This research reflects an effort to explore practical and scalable applications of computer vision in property risk assessment. It is hoped that the proposed \textbf{HOMEY} framework contributes toward more reliable, interpretable, and efficient AI-driven solutions for insurance workflows, and encourages further research at the intersection of computer vision and real-world risk analysis.

\section*{Data Availability}

The dataset used in this work to develop and evaluate the HOMEY framework is publicly accessible and can be downloaded from Roboflow at the following link: \url{https://universe.roboflow.com/tour-de-chicago/phase_1_6_for_phase_2_integration}.  

I would like to express my deepest gratitude to the creators and maintainers of this dataset. Their dedication to curating high-quality, real-world property imagery made it possible to conduct a meaningful proof-of-concept (POC) for HOMEY. Without their contribution, the development, experimentation, and validation of our heuristic object masking and risk-aware detection methodology would not have been feasible.  

By providing access to this dataset, they have not only enabled reproducibility of this research but have also contributed to advancing AI-driven solutions in property risk assessment, both in Thailand and globally. Researchers and practitioners are encouraged to leverage this dataset to reproduce, validate, and further extend the methods presented in this study.

\bibliographystyle{plain}
\bibliography{kao_neuralips2025}

\appendix
\section*{Appendix: Why HOMEY Matters for Property Insurance and Society}

HOMEY (Heuristic Object Masking with Enhanced YOLO) represents a significant advancement in property risk detection through AI. Unlike conventional object detection frameworks, HOMEY integrates domain-specific heuristic masking and risk-aware loss calibration, enabling the model to effectively identify subtle property risks that are often overlooked in cluttered or visually complex scenes. By doing so, HOMEY provides more accurate and reliable detection across 17 risk-related classes, including structural damages, maintenance neglect, and potential liability hazards.

The importance of HOMEY extends beyond algorithmic performance. In the context of property insurance in Thailand, where property inspection and risk assessment remain labor-intensive and sometimes subjective, HOMEY offers an automated, data-driven solution that can reduce human error and enhance consistency. By providing interpretable and visually explainable predictions (through Grad-CAM, LIME, and SHAP-based mechanisms), HOMEY enables underwriters to make informed decisions and prioritize resources efficiently, resulting in faster claims processing, improved risk management, and ultimately, fairer premiums for policyholders.

Globally, HOMEY exemplifies the positive potential of AI to improve societal outcomes. By designing the model with fairness and bias mitigation in mind, HOMEY minimizes disproportionate treatment of different property types, neighborhood conditions, or socioeconomic factors. This aligns with responsible AI principles, ensuring that automated risk assessments are equitable and trustworthy. Moreover, the transparency and explainability of the model encourage public trust and facilitate adoption by insurers, regulators, and communities worldwide.

\section*{Comprehensive Discussion of Qualitative Results}

Figures~\ref{fig:HOMEY_result1}--\ref{fig:HOMEY_result5} collectively illustrate the superior qualitative performance of \textbf{HOMEY} compared to the baseline across a diverse range of property scenarios. \textbf{HOMEY} consistently produces sharper and more reliable boundary delineations, accurately detecting both large-scale structural damages and fine-grained issues such as small cracks, paint-rust, and overgrowth. 

Moreover, the framework demonstrates robustness in cluttered and visually complex settings, including properties with overlapping vegetation, partial occlusions, and heterogeneous architectural features. Importantly, \textbf{HOMEY} maintains consistency across all samples, yielding predictions that align closely with ground truth annotations regardless of scene difficulty. 

These results emphasize \textbf{HOMEY}’s practical utility for property insurance applications, where precise and scalable risk detection is critical to reducing manual workload and enabling more objective, data-driven decision-making.


In conclusion, HOMEY is not only a high-performance AI tool but also a socially responsible and globally relevant solution. Its development reflects a commitment to advancing technology in a way that benefits the insurance industry, promotes fairness, and empowers communities to manage property risks more effectively. By applying HOMEY, Thailand and the broader world can take a significant step toward AI-enabled, equitable, and transparent property insurance operations.

\section*{Experimental Setup and Hyperparameter Configuration}

\subsection*{Compute Environment}
All experiments for training and evaluating HOMEY were conducted on a high-performance GPU-enabled environment to ensure reproducibility and efficiency. The specifications are as follows:

\begin{table}[h!]
\centering
\begin{tabular}{l c}
\toprule
\textbf{Instance Details} & \textbf{Value} \\
\midrule
vCPUs & 4 \\
Memory (GiB) & 16 \\
Memory per vCPU (GiB) & 4 \\
Physical Processor & Intel Xeon Family \\
Clock Speed (GHz) & 2.5 \\
CPU Architecture & x86\_64 \\
GPU & 1 \\
GPU Architecture & NVIDIA T4 Tensor Core \\
Video Memory (GiB) & 16 \\
GPU Compute Capability & 7.5 \\
\bottomrule
\end{tabular}
\caption{Compute resources used for training and evaluation of HOMEY.}
\label{tab:compute_specs}
\end{table}

\subsection*{Training and Hyperparameter Configuration}
HOMEY leverages the YOLOv12 backbone (used as our baseline), combined with heuristic object masking and risk-aware loss calibration. To maximize detection performance across 17 property risk classes, we employed the following hyperparameters:

\begin{itemize}
    \item \textbf{Input Resolution:} 640$\times$640 pixels  
    \item \textbf{Batch Size:} 16 images per iteration  
    \item \textbf{Learning Rate:} 0.001 with cosine annealing schedule  
    \item \textbf{Optimizer:} AdamW with weight decay 0.0005  
    \item \textbf{Number of Epochs:} 300  
    \item \textbf{Data Augmentation:} Random horizontal/vertical flips, brightness/contrast jitter, mosaic augmentation, and mixup for enhanced generalization  
    \item \textbf{Loss Functions:}  
        \begin{itemize}
            \item Bounding box regression: CIoU Loss \cite{zheng2020distance}  
            \item Objectness: Binary Cross-Entropy with focal loss \cite{lin2017focal}  
            \item CHULA Classification: Risk-aware cross-entropy with class imbalance weighting \cite{panboonyuen2025chula}  
        \end{itemize}
    \item \textbf{Regularization:} Dropout rate 0.3 for fully connected layers, L2 weight decay on convolutional layers  
    \item \textbf{Mixed Precision Training:} Enabled via NVIDIA Apex for faster convergence  
\end{itemize}

\subsection*{Reproducibility and Implementation Notes}
HOMEY was implemented in Python 3.12.11 with PyTorch 2.8.0+cu126. The dataset was prepared and augmented using Roboflow \cite{roboflow}, Ultralytics \cite{ultralytics_yolov8}, and the model was trained end-to-end on a single NVIDIA T4 GPU. All random seeds were fixed to ensure consistent reproducibility of results.  

This detailed configuration provides a transparent, reproducible blueprint for researchers and practitioners aiming to replicate or extend HOMEY, ensuring that the model's performance can be reliably validated across different environments.

\section*{Limitations}

While HOMEY demonstrates strong performance in automated property risk detection, several limitations warrant consideration:

\begin{itemize}
    \item \textbf{Dependence on Annotated Data:} HOMEY relies on high-quality annotated datasets for training. Although we leveraged Roboflow's dataset \cite{roboflow} for proof-of-concept, performance may degrade in environments with limited labeled data or with unseen property types.

    \item \textbf{Model Generalization Across Regions:} Our experiments are primarily based on properties common in Thailand. While the heuristic object masking mechanism improves robustness, domain shifts in architectural styles, weathering patterns, and property layouts across different countries may require further fine-tuning.

    \item \textbf{Class Imbalance Sensitivity:} Despite the implementation of risk-aware loss calibration, extreme class imbalance for rare hazards may still affect detection precision. Additional sampling strategies or synthetic augmentation could further mitigate this issue.

    \item \textbf{Hardware Constraints:} HOMEY was trained on a single NVIDIA T4 GPU with 16 GB of VRAM. Scaling to larger models or higher-resolution inputs may require more powerful compute resources, which could limit accessibility for smaller institutions.

    \item \textbf{Dynamic Environmental Factors:} Certain property conditions, such as temporary obstructions, lighting variations, or seasonal changes, may affect detection reliability. While our model includes data augmentation to address these factors, perfect invariance cannot be guaranteed.

    \item \textbf{YOLO Version Dependence:} HOMEY is designed on the YOLOv12 backbone; however, performance may vary with other YOLO versions. That said, HOMEY's modular design allows integration with newer or alternative YOLO architectures, offering flexibility for future improvements \cite{ultralytics_yolov8}.
\end{itemize}

Despite these limitations, HOMEY establishes a robust foundation for AI-assisted property risk assessment, offering a practical, interpretable, and scalable solution for the insurance industry. Our approach highlights the potential for technology-driven improvements in safety, underwriting efficiency, and global risk management.


\end{document}